\documentclass[letterpaper, 10 pt, conference]{ieeeconf}  % Comment this line out if you need a4paper
\usepackage{graphicx}

\usepackage{amssymb}
\usepackage{graphics}
\usepackage[backend=biber, style=ieee]{biblatex}
\usepackage{multirow}
\usepackage{pifont}
\usepackage{hyperref}
\usepackage{booktabs}
\usepackage{amsmath}
\DeclareMathOperator*{\argmax}{argmax}

\IEEEoverridecommandlockouts                              % This command is only needed if 
\title{\LARGE \bf
MBAPPE: MCTS-Built-Around Prediction for Planning Explicitly
}

\author{Raphael Chekroun$^{*,1,2,3}$, Thomas Gilles$^{*,1}$\\ Marin Toromanoff$^2$, Sascha Hornauer$^1$, Fabien Moutarde$^1$ 
\thanks{$^{1}$ Mines Paris - PSL University, Centre for Robotics, Paris, France}
\thanks{$^{2}$Valeo Driving Assistance Research, Créteil, France}
\thanks{$^{3}$Department of Civil and Environmental Engineering, University of California, Berkeley, USA}
\thanks{$^{*}$Equal contributions}}

\begin{document}

\maketitle
\thispagestyle{empty}
\pagestyle{empty}

%%%%%%%%%%%%%%%%%%%%%%%%%%%%%%%%%%%%%%%%%%%%%%%%%%%%%%%%%%%%%%%%%%%%%%%%%%%%%%%%
\begin{abstract}
We present MBAPPE, a novel approach to motion planning for autonomous driving combining tree search with a partially-learned model of the environment. Leveraging the inherent explainable exploration and optimization capabilities of the Monte-Carlo Search Tree (MCTS), our method addresses complex decision-making in a dynamic environment. We propose a framework that combines MCTS with supervised learning, enabling the autonomous vehicle to effectively navigate through diverse scenarios. Experimental results demonstrate the effectiveness and adaptability of our approach, showcasing improved real-time decision-making and collision avoidance. This paper contributes to the field by providing a robust solution for motion planning in autonomous driving systems, enhancing their explainability and reliability.
Code is available under \url{https://github.com/raphychek/mbappe-nuplan}.
\end{abstract}

\section{INTRODUCTION}
%The increasing demand for safe and efficient autonomous navigation necessitates advanced planning techniques. 
Innovations in machine learning techniques have led to significant advancements in self-driving technology. Particularly, the use of deep learning has greatly improved the perception stage of autonomous driving. These developments have been complemented by progress in sensor technology and mapping methods. As a result, the focus is now shifting to the next challenges of autonomous driving, and motion planning emerges as a pivotal component. After identifying roads and monitoring nearby vehicles and object entities, the autonomous driving system must now decide its future path and plan its trajectory accordingly to ensure a collision-free route while respecting traffic rules. 

Therefore, this study centers on the mid-to-end stage of autonomous driving, presuming that perception tasks have already been accomplished and working toward an efficient and explainable motion planning. In this realm, recent research mostly focus on Imitation Learning (IL) \cite{gohome, gameformer, plancnn} or hybrid IL and rule-based methods \cite{pdm, gc-pgp}.

However, rule-based methods for autonomous driving are limited by their lack of scalability, adaptability, robustness in complex and ambiguous situations, and their inability to handle unconventional scenarios. This contrasts with machine-learning based approaches that address these limitations through data-driven learning and adaptability. 

\begin{figure}
    \centering
    \includegraphics[width=\columnwidth]{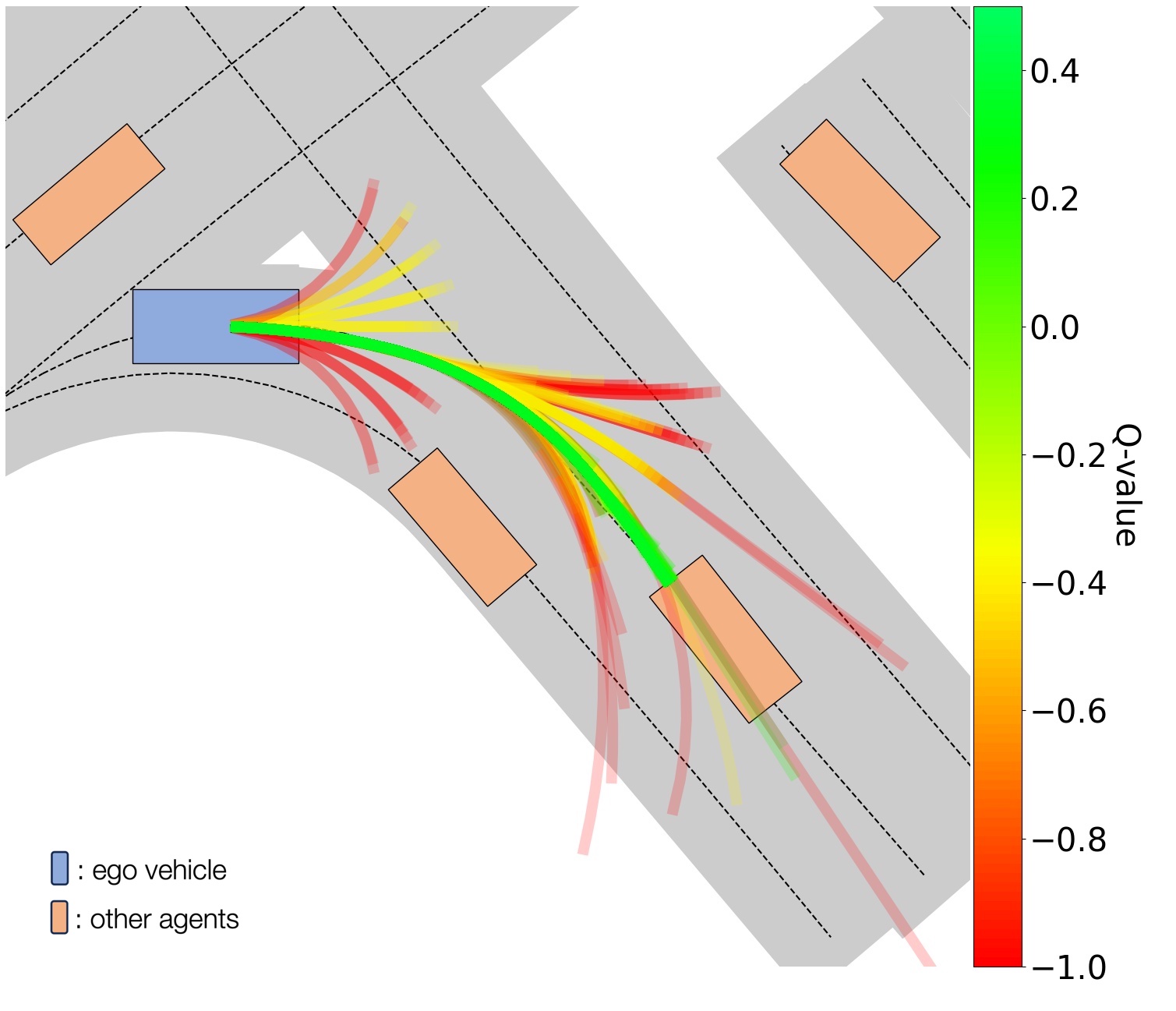}
    \caption{Visualization of the exploration done by MBAPPE in one planning step. We display the bird-eye-view trajectory pieces in xy coordinates. As the road is turning right, the MCTS explores multiple steering angle and acceleration configurations to correctly take the turn. MBAPPE finally selects the path which maximizes the Q-value (in green).}
    \label{fig:exploration}
\end{figure}

Nonetheless, while Neural Networks (NN) provide a powerful and flexible tool for learning to drive using supervised labels with IL methods \cite{bojarski,transfuser,gohome}, they remain limited in the long-term understanding of the consequences of their actions. Therefore, they may not comprehend the full scope of interactions with the map and other agents. Deep Reinforcement Learning (Deep RL) based methods \cite{day, wor, griad} aim to incorporate long-term returns of such consequences in the training of these networks. However, this causal understanding remains implicit and not guaranteed, and Deep RL training is most often sample inefficient. 

\begin{figure*}[t!]
    \centering
    \includegraphics[width=\linewidth]{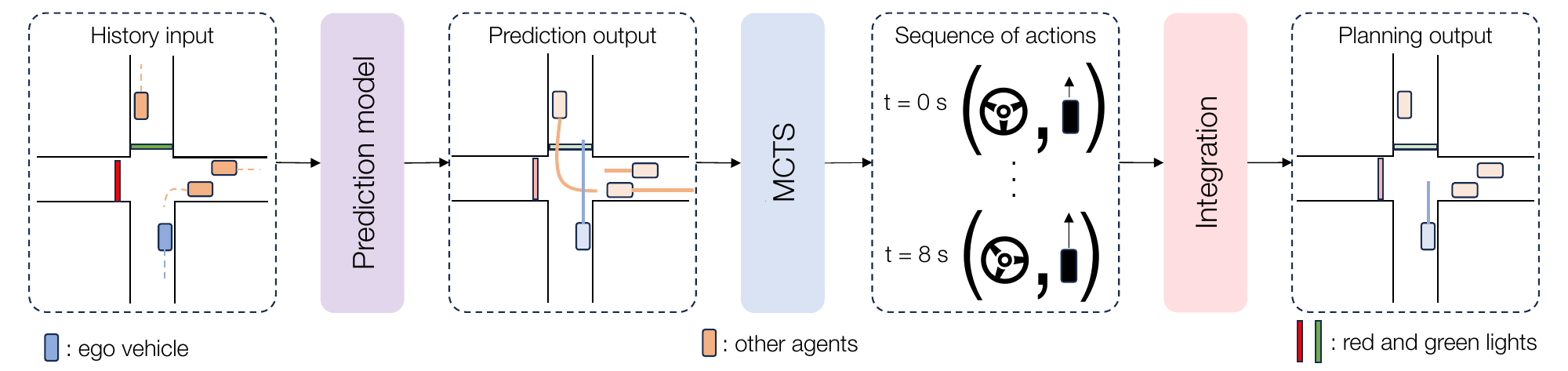}
    \caption{\textbf{MBAPPE pipeline} A prediction model infers future trajectories of other agents in the scene. This information is fed to the MCTS which outputs a sequence of consecutive actions. Those are integrated to form an improved trajectory planning for the ego.}
    \label{Fig:mbappe_fig}
\end{figure*} 

Our approach aims to get the best of both worlds by using an IL prior to guide a MCTS \cite{mcts_1, mcts_2} into explicitly exploring the consequences of actions, validating the NN trajectory if it respects driving constraints, or exploring new actions if required, see Figure \ref{fig:exploration}. The main challenge in running a MCTS is that it assumes environment transitions to be deterministic and perfectly known. While this is true for the displacement of the ego vehicle given its actions, and for the update of the map that remains the same, other agents will also move on their own accord. In order to have a realistic world model, we developed an IL model to predict all the other agents future trajectories. This way we get an approximate of the future transitions that enables us to roll out the consequences of our chosen actions on multiple time-steps. 

In this paper, we extend the MCTS paradigm to partially-learned environment and apply it to autonomous driving. Next, we validate our performance on nuPlan \cite{nuplan} simulation environment and compare to other existing baselines. Lastly, we highlight the explainability of our approach which allows easy observation and analysis of the steps leading to any given decision via its decision tree.

\section{Related work}
MBAPPE seeks to leverage imitation learning (IL) to guide a MCTS model in exploring the outcomes of its actions. As such, this section is dedicated to examining rule-based and learning-based motion planning techniques, and strategies integrating MCTS with deep learning.

\paragraph{Rule-based methods}
Rule-based methods employ explicit rules to dictate the behavior of autonomous vehicle, making them interpretable by nature \cite{stanley, rule_based, odin}. A notable instance is the Intelligent Driver Model (IDM) \cite{idm}, designed to track leading vehicles while maintaining safe distances through computation of optimal acceleration based on the leading vehicle's speed. Rule-based methods were extended in predictive rule-based approaches which anticipate future environmental states to improve collision avoidance \cite{pavone, leader, neural_motion_planner}. 
However, rule-based methods are inflexible and rely on perfect and consistent representation of the environment. This characteristic make them struggle with generalization to novel scenarios or with the inherent variability of real-world conditions.

\paragraph{Imitation learning methods}
%Overcoming the need of a perfect representation per its generalizability, imitation learning methods allow to learn driving policies from supervised data. Regarding implicit deep learning based methods, a network directly generates planned trajectories or control commands \cite{urbandriver, gameformer, transfuser}. However, such approaches suffers a lack of interpretability and general robustness.

%To tackle those issues, other methods aim at more interpretability. Dauner et al.'s Predictive Driver Model (PDM) \cite{pdm} aims at complementing an IDM, which is interpretable by nature, with a light neural networks \cite{pdm}. Some methods mitigate the lack of robustness of implicit methods by generating several deep based trajectories and selecting the optimal candidate presenting the lowest cost \cite{lookout, perceive, neural, dsdnet}. Huang et al.'s GameFormer  \cite{gameformer} implicitly generates a single trajectory which is then refined with a non linear optimizer.

Imitation learning methods allow to learn how to drive from supervised data, leading to more generalizability than rule-based methods. Some of these methods directly create driving plans or commands \cite{urbandriver, transfuser}, but they suffer from a lack of interpretability and general robustness. To address these issues, some other approaches focus on making the planning decisions more interpretable. For instance, Dauner et al. developed Predictive Driver Model (PDM) \cite{pdm} to combine an interpretable IDM with a simple neural network. Some methods deal with the robustness problem by generating multiple planning options with deep learning and then choosing the best one with the lowest cost \cite{lookout, perceive, neural, dsdnet} or by refining deep-based predictions \cite{gameformer, adapt}. 
However, IL methods still suffers from distribution mismatch where agent fails to recover from accumulation error thus leading to increasingly out of expert distribution states, and lacks of long-time reasoning.

% Scheel et al.'s Urban Driver open-loop \cite{urbandriver} leverages graph attention on agents and map features vectorized similarly than in Gao et al.'s VectorNet \cite{vectornet} for input processing and infers a trajectory via another multi-head attention layer. 

% At the interface of rule-based methods and imitation learning, Dauner et al.'s Predictive Driver Model (PDM) method implements a simple MLP which inputs are centerline features extracted with an extended IDM and vectorized agent history. %While simple, this method reaches outstanding results on nuPlan and was ranked \#1 on the first edition of the nuPlan challenge. 

% %Hu et al.'s hoplan method  \cite{hoplan} trains a CNN to build a bird-eye-view heatmap \cite{home}, an occupancy map from six-channel raster representation of the scene and a trajectory estimation. This estimation is then refined using a CasADi \cite{casadi} ipopt solver.

% Huang et al.'s GameFormer Planner \cite{gameformer-planner} exploits the GameFormer model \cite{gameformer}, a Tranformer-based network \cite{transformer}, to compute ego and other agents trajectory estimations from scene features and a rule-based obtained reference path. This trajectory estimation is then refined with a nonlinear optimizer to mitigate IL distributional shifts and causal confusion.

\paragraph{Reinforcement learning methods}
%Reinforcement learning paradigm aims at a lessen dependence on expert data and an improved ability to handle more varied tasks. Indeed, leveraging a reward function to assess the quality of a policy instead of mimicking human behavior like IL does allows RL models to reach efficient behavior, sometimes even outperforming humans \cite{alphago}. 
%Reinforcement learning aims to rely less on expert data and get better at handling different tasks with more variety. 
Instead of copying human behavior like IL, RL models use a reward system to judge how good a strategy is. This can lead to improved decision-making, sometimes even outperforming humans \cite{alphago}. Model-free reinforcement learning focuses on learning optimal actions directly from observed states and rewards without creating an explicit model of the driving environment. Even though RL is successful for simple autonomous driving tasks \cite{day}, up to now, no published work has reported sucess of exclusively RL-based method in autonomous driving for complex urban environments \cite{bernhard}. Furthermore, RL suffers from sample inefficiency and lack of convergence guarantees and interpretability. Recent works leveraged supervised learning in RL pipelines to overcome these limitations \cite{toro, griad}, thus compensating the weakness of the RL gradient during training. 
% Aiming at greater adaptability to new scenarios, less dependency on expert data, and the ability to handle a wider variety of tasks than IL which mainly focuses on mimicking human behavior, the use of RL is becoming more widespread in the field of autonomous driving.

%Model-free reinforcement learning learns to make decisions directly from its experiences and rewards, without constructing a model of the environment.  In contrast, model-based reinforcement learning involves building an explicit model of the environment, which the agent uses to plan and make decisions. Model-free methods often include algorithms like Q-learning, while model-based approaches use planning algorithms to optimize behavior.

\begin{figure*}
    %\centering
    \includegraphics[width=\linewidth]{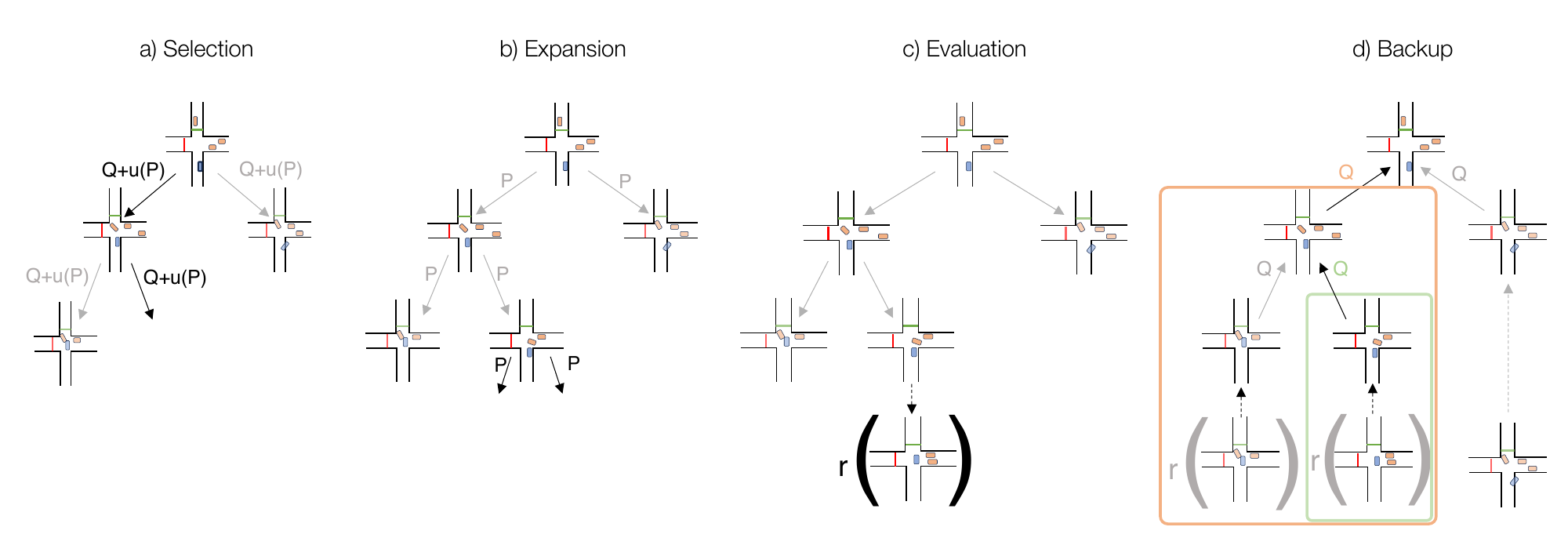}
    \caption{\textbf{MCTS steps} a) Each simulation pass in the tree follows a trade-off between exploitation of the best $Q$ value of an action, and the exploration term $u(P)$ that encourages to explore nodes with less visits $N$ along the prior $P$. b) The leaf node is possibly expanded following some probabilities depending on the prior $P$ and the continuity constraints. c) After the simulation, the leaf node is evaluated by explicitly computing the reward $r$ described in Section \ref{sec:mcts_design}. d) $Q$-values are updated so means of the rewards $r$ in the sub-tree below each actions are tracked.}
    \label{Fig:mcts_fig}
\end{figure*}

\paragraph{Methods integrating MCTS with deep learning}

Integrating MCTS with deep learning techniques has emerged as a compelling approach to enhance decision-making processes in various domains. Silver et al. \cite{alphago} pioneered this fusion by combining MCTS with deep supervised learning to achieve groundbreaking results in the game of Go with AlphaGo. This paradigm was extended with AlphaZero \cite{alphazero} by relying solely on self-play and RL. MuZero \cite{muzero} finally embraced implicitness and extended the generality of these approaches by employing learned models to simulate outcomes and inform strategic decision-making.

In the realm of autonomous driving, Chen et al. \cite{chen} integrated MCTS with deep learning but relied on implicitness for the tree transitions and prior computation, possibly leading to inexplicable behaviors which are not desirable for this domain of application. Other published methods lack generalizability and constraint their applicative fields to simplified custom environments such as highway driving without possibility for public benchmarks comparison \cite{mcts_restrictive_3, mcts_restrictive_2}, or high level tactical decisions \cite{mcts_restrictive}.

%Chen et al. \cite{chen} proposed a MuZero-like where root state are encoded via a CNN and state-to-state transitions, actions and value are inferred by NNs. While promising, this implicitness leads to a lack on interpretability which is not desirable for this domain of application. Ha et al. \cite{mcts_restrictive_2} also proposed a method with implicit learned-based value and reward on the restrictive task of highway driving. On the same task, Hoel et al. \cite{mcts_restrictive} extended these methods to continuous action-space for tactical decision making, leaving the final control to an IDM.

%General:
%- alphazero, alphago, muzero
%
%Autonomous driving 
%- https://arxiv.org/pdf/1905.02680.pdf : easy scenario
%- https://github.com/chauvinSimon/My_Bibliography_for_Research_on_Autonomous_Driving/blob/master/sections/8_planning_mcts.md

\section{Method}

In this section, we introduce MBAPPE and its components. In particular, we present the known and learned features of the world model, and technical details of our MCTS design and exploration steps.

\subsection{MBAPPE framework}
At each time-step, a neural network (based on an open-loop version of Urban Driver \cite{urbandriver}) predicts an estimation of the ego trajectory and of the future trajectories of every other agents around the ego. This information is fed to the MCTS, which will deploy an internal lightweight simulation where the ego trajectory is used as a prior to guide the first steps of exploration, and other agents trajectories are leveraged to build the world model. At each simulation-step, which follow a planning time axis inside the tree, the MCTS explores the possible actions and internally simulates the evolution of the environment to check how those explored actions will impact its driving performances (driving out of area, check for collisions with static objects, check collisions with other agents thanks to their estimated trajectory, etc). 

The global pipeline is represented in Figure \ref{Fig:mbappe_fig}.

\subsection{World Model}
The Monte-Carlo tree search leverages an internal simplified representation of the world where it can quickly iterate to explore possible sequences of actions and their consequences.
This environment is made of two categories of features:
\begin{itemize}
    \item Known features: 
        \begin{itemize}
            \item The map information, including traffic light, 
            \item Static objects such as traffic cones and barriers
            \item Dynamic objects such as neighboring vehicles, traffic cones or pedestrians, which we will consider as other agents evolving in the simulated environment
        \end{itemize}
    \item Learned features: 
    \begin{itemize}
        \item Estimated future trajectories of other agents given by the NN prediction.
    \end{itemize}
\end{itemize}

\subsection{MCTS design and tree steps}
\label{sec:mcts_design}
Our MCTS is based on a kinematic bicycle model of the vehicle. Actions are defined as a tuple $(a, \delta)$, where $a$ is the acceleration and $\delta$ the steering angle. Accelerations and steering are discretized in 13 values each, in the respective range of $[-3, 3]$ m.s$^{-2}$ and $[-\pi/4, \pi/4]$ rad. Actions are integrated every 0.1 s. 

The simulation process of our tree search is detailed in Fig. \ref{fig:exploration}. The tree is initialized with a single root node representing the current context.
Each tree node stores 3 values: Q the expected return, P the action prior and N the number of visits. The nodes are built and evaluated iteratively through the following steps:

\begin{itemize}
    \item \textbf{Selection}: We follow the PUCTS \cite{alphago} formula to select the next action following a trade-off between the exploitation of Q and the exploration of unvisited nodes with low $N$. 
    
    At a node state $\mathcal{S}$ the action $\mathcal{A}$ is chosen using the following formula:
        \begin{equation}
            \mathcal{A}_t = \argmax_{\mathcal{A}} Q(\mathcal{S}, \mathcal{A}) + c_{puct}P(\mathcal{S},\mathcal{A})\frac{\sqrt{\sum_\mathcal{B} N(\mathcal{S},\mathcal{B})}}{1+N(\mathcal{S},\mathcal{A})}
        \end{equation}
    with $c_{puct}$ an hyper-parameter balancing the trade-off between exploration and exploitation. We found $c_{puct}=2$ to perform the best in our experiments.
    \item \textbf{Expansion}: We expand leaf nodes by all physically possible actions from the state of the leaf node, following a prior $P$ and some continuity constraints. These constraints ensure both comfort and physical feasibility of successive actions. Prior design and continuity constraints are described in Section \ref{sec:prior}. %The prior P and additional continuity constraints, added to have successive actions be close to each other and respect realistic physics, are described in the section \ref{sec:prior}.  
    %The prior P is given by:
    %    \begin{itemize}
    %        \item If the leaf node is less than a certain depth: the sum of a Gaussian centered of the action derived from the NN ego prediction and a Gaussian centered on constant speed and steering,
    %        \item Else: a Gaussian centered on constant speed and steering
    %    \end{itemize}
    %    Additional continuity constraints are added to have successive actions be close to each other and respect realistic physics.
    \item \textbf{Evaluation}: We consider that driving rewards are rather short term (crash or not, exit road or not within the next 6 or 8 seconds). Therefore they do not need to be bootstrapped by a learned value network, but rather can be evaluated at the current simulation step by checking for them directly. Our computed reward $r_t$ at state $s_t$ is made of these main components:
        \begin{itemize}
            \item Progress: distance advanced since the last node, normalized by maximum allowed speed limit ($[0,1]$),
            \item Collision: penalty for collision with car and pedestrian ($-5$) or object ($-2$),
            \item Route: $-0.5$ if the vehicle is not on the expected road,
            \item Drivable area: $-1$ if the vehicle is not on the drivable area,
            \item Center of the road: 
                \begin{itemize}
                    \item $-sin(\theta)/2$ where $\theta$ is the angle difference between the ego heading and the closest centerline heading,
                    \item $-d/2$ where $d$ is the distance between the ego position and the closest centerline.
                \end{itemize}
        \end{itemize}
    \item \textbf{Back up}: We update the Q values using the cumulative reward as in MuZero \cite{muzero}:
        \begin{equation}
        \begin{gathered}
            G^k=\sum_{\tau=0}^{l-1-k} \gamma^\tau r_{k+1+\tau} \\
            Q\left(s^{k-1}, a^k\right):=\frac{N\left(s^{k-1}, a^k\right) \times Q\left(s^{k-1}, a^k\right)+G^k}{N\left(s^{k-1}, a^k\right)+1} \\
            N\left(s^{k-1}, a^k\right):=N\left(s^{k-1}, a^k\right)+1 
        \end{gathered}
        \end{equation}
    We use a discount factor $\gamma$ of 1.
    
\end{itemize}

%Since our computing time bottleneck is in the evaluation part, and more precisely in the checking for collisions and drivable area, we do not evaluate at every timestep to speed up the MCTS process time. Instead we only evaluate nodes every 10 nodes (every 1 second)  and back-propagate their values in the previous un-evaluated nodes.

\subsection{Prior and continuity constraints}
\label{sec:prior}

An efficient MCTS exploration process can be achieved by leveraging two approaches. %via two vectors. 

Firstly, providing the MCTS an intuition over actions to explore to prioritize the more probable ones. This issue is tackled using a prior over the distribution of actions for each node. This prior is usually learned and inferred for every node \cite{muzero}, which is computationally expensive, or handcrafted. Secondly, to further streamline the exploration process, we narrowed down the action space, thereby reducing the overall actions that need to be explored to the most critical ones. To achieve this, we integrated continuity constraints into the MCTS to ensure not only the physical feasibility of the actions explored but also to enhance comfort and to reduce the exploration time.

\subsubsection{The prior}
We designed a prior which relies on both handcrafted rules and learned rules, all without incurring any additional computational overhead. % but without any supplementary computational time required.

The prior function is made of two parts:
\begin{itemize}
    \item The handcrafted prior $P_h$ prioritizes exploration around the constant speed with null steering angle,
    \item The learned prior $P_l$ is obtained by deriving the prediction of the ego trajectory by the NN into consecutive actions. This prior advantages the possibility of following NN actions for the first $T$ time steps of the internal simulation of the MCTS. We found $T=1$ s to perform the best in our experiments.
\end{itemize}

Both $P_l$ and $P_h$ are Gaussians centered on the chosen action. The Gaussian are parametrized with a very high variance ($\sigma^2=100$) to encourage an almost uniform exploration.

The designed prior can be written:

\begin{equation}
  P^t =
    \begin{cases}
      P_h^t + P_l^t & \text{if $t \leq T$}\\
      P_h^t &  \text{if $t > T$}
    \end{cases}       
\end{equation}

\subsubsection{Continuity constraints}

To ensure the output trajectory is physically feasible and to minimize the total number of actions to explore, we implemented continuity constraints in the MCTS.

These constraints are two folded:
\begin{itemize}
    \item The Tree Constraint: At a given step $t$ of the real-world vehicle movement, the root node of the novel tree will be constrained to explore neighboring accelerations and steering angles relatively to the actions taken at time $t-1$ by the previous tree. This constraint favors a behavior continuity between successive time-steps and corresponding MCTS.
    \item The Node Constraint: During the MCTS internal expansion phase, exploration only focuses on neighboring accelerations and steering angle values relatively to the actions of his parent node. This constraint favors a behavior continuity during the expansion phase of a given MCTS.
\end{itemize}

We formulate both continuity constraints as restricting the following action $(a_{t+1}, \delta_{t+1})$ to be within a range of $a_{t} \pm 0.15$ m.s$^{-2}$ for the acceleration and $\delta_{t} \pm \pi/240$ rad for the steering angle with $(a_{t}, \delta_{t})$ the action at the previous time-step.

\section{Experimental results}

%Our focus rests on the nuPlan dataset. 
\textbf{Dataset: }
We show results on the nuPlan dataset. It encompasses 1300 hours worth of real vehicle motion data along with its corresponding simulator.  Within the nuPlan framework, we chose to assess the performance of planners on closed-loop non-reactive agents benchmark. 
We focus on this benchmark, as evaluations conducted in closed-loop more effectively assess an agent's driving capabilities without the need to compare them to a flawed 'ideal' behavior as typically seen in open-loop assessments. Additionally, we chose non-reactive agents for our study, as preliminary experiments and other performance benchmarks \cite{nuplan, pdm} have demonstrated that outcomes are largely consistent between reactive and non-reactive agents. All simulations are ran on 100 scenarios of each of the 14 scenarios types (totaling 1,118 scenarios in practice, as all 14 types do not have 100 available scenarios) of the nuPlan challenge, following the Val14 benchmark validation set \cite{pdm}.  

\textbf{Score and metrics: }
We use the nuPlan official score, which measures driving quality between 0 and 100 through a combination of 16 normalized driving metrics related to infraction rate, ego comfort, or progress toward the goal.
 We decided to put a special emphasis on the metrics of collision rate (CR), driving area non-compliance (DA) and ego progress (EP) in our experiments, as they are key elements for a safe and efficient autonomous driving system.

%All models have been trained on a maximum of $4k$ of each type of nuPlan scenarios, giving a total of $\sim 177k$ training scenarios. 

\textbf{Implementation details: }
For ablations studies, the number of simulation steps is limited to 256 in each MCTS. In our setup (Intel Core i7-9700K CPU @ 3.60GHz) the whole pipeline inference time is $\sim 0.15$ seconds for this setup, including input pre-processing, prediction model, MCTS and post-processing. The pipeline runs on CPU only. 
For inference speed purposes, we only expand new possible actions every 1 s. We observed no drop of performance.

%Notably, the MCTS leveraging the NN prior performs only slightly better than the MCTS without any learned NN prior (each node is just initialized with a very flattened Gaussian centered on null acceleration and straight steering). This can be explained by the poor performance of the NN alone, very far from the reported baseline performance of the challenge. We also believe that we have not explored enough the different ways in which the NN prior could be used.

\subsection{Ablation study over the prior}

An ablation study over the choice of prior is presented table \ref{tab:ablation_prior}. Continuity constraints are the one described section \ref{sec:prior}.

\begin{table}[!ht]
    \centering
    \resizebox{\columnwidth}{!}{
    \begin{tabular}{cc|ccc|c}
    \toprule
    \multicolumn{2}{c|}{Prior} & \multicolumn{3}{c|}{Metrics} \\
    Learned & Crafted & CR $\downarrow$ & DA $\downarrow$ & EP $\uparrow$ & Score $\uparrow$ \\ 
    \midrule 
    - & - & 6\% & 4\% & 31\% & 26\% \\
    \ding{51} & - & 11\% & 3\% & 88\% & 65\% \\
    - & \ding{51} & 6\% & 4\% & 95\% & 82\% \\
    \ding{51} & \ding{51} & \textbf{5\%} & \textbf{2\%} & \textbf{96\%} & \textbf{86\%} \\
    \bottomrule
\end{tabular}
}
    \caption{Ablation over the prior.} 
    % We observe that incorporating both learned and crafted priors significantly enhances performance metrics.}We observe that using both prior drastically increases scores. In particular the learned prior guides the MCTS to stay inside the driving area, and the crafted prior reduces the number of collision and improves overall progress. Combining learned and crafted prior led to best results overall.}
    \label{tab:ablation_prior}
\end{table}

We can see from results of Table \ref{tab:ablation_prior} that MCTS without prior is inefficient. Exploration being unguided, the expansion phase does not create node leading to a good reward \textit{a priori}. Following $P_l$ for the first steps of the simulation allowed to significantly improve the exploration phase by guiding the MCTS to stay within the driving area. Indeed, thanks to continuity constraints, a good beginning of the trajectory allows to stay on the road and reach acceptable metrics.  Interestingly, leveraging only $P_l$ leads to an increase of the collision rate: if the MCTS first actions differ from the prior's, there will be a mismatch between the guidance it provides and the actual scenes which can lead to collisions.   %this learned prior alone allows the MCTS to outperform Urban Driver initial prediction (cf. Table \ref{tab:val14}). 

Leveraging $P_h$ allows the MCTS to prioritize exploration of the most common behavior on average (staying at around the same velocity with a null steering angle), therefore minimizing collisions and optimizing overall progress. Notably, using only this naive prior without any kind of learning already yields very good performance, highlighting the power of guided exploration in the MBAPPE method. Finally, leveraging $P_h + P_l$ allows to prioritize this kind of behavior while starting with a better heads up and leads to best results on this set of experiments.

\subsection{Ablation study over continuity constraints}

An ablation study over the choice of continuity constraints is presented Table \ref{tab:ablation_constraint}. For these experiments, prior is $P_h + P_l$.

\begin{table}[!ht]
    \centering
    \vspace{0.2 cm}
    \resizebox{\columnwidth}{!}{
    \begin{tabular}{cc |ccc|c}
    \toprule
    \multicolumn{2}{c|}{Constraints} & \multicolumn{3}{c|}{Metrics} & \\
    Tree & Node & CR $\downarrow$ & DA $\downarrow$ & EP $\uparrow$ & Score $\uparrow$ \\ 
    \midrule 
    - & - & 7\% & 4\% & 95\% & 79\% \\
    \ding{51} & - & 8\% & 3\% & 96\% & 82\% \\
    - & \ding{51} & 8\% & 2\% & 94\% & 82\% \\
    \ding{51} & \ding{51} & \textbf{5\%} & \textbf{2\%} & \textbf{96\%} & \textbf{86\%} \\
    \bottomrule
\end{tabular}
    }
    \caption{Ablation over continuity constraints.}
    \label{tab:ablation_constraint}
\end{table}

%\begin{table}[H]
%\centering
%\begin{tabular}{|c|l|r|r|r|r|}
%\hline
%& \multicolumn{1}{c|}{Text} & \multicolumn{1}{c|}{Text} & \multicolumn{1}{c|}{Text} & \multicolumn{1}%{c|}{Text} & \multicolumn{1}{c|}{text}\\%
%\hline
%\parbox[t]{2mm}{\multirow{3}{*}{\rotatebox[origin=c]{90}{rota}}} & text &&&&\\
%& text &&&&\\
%& text &&&&\\
%\hline
%\end{tabular}
%\end{table}

It becomes apparent that when applied separately, continuity constraints offer only marginal improvements to our method. A possible explanation is that the handcrafted identity prior already directs the MCTS towards a form of constrained exploration similar to what is achieved through node constraints. However, utilizing both node and tree constraints independently does enhance the exploration process. Importantly, the combined effects of these constraints not only substantially increase performance but also ensure a consistent selection of actions, both within a single tree and across multiple trees that correspond to sequential planning steps.

\subsection{Comparison with state-of-the-art methods}

%We compare MBAPPE's performance with other state-of-the-art method on a benchmark made of 100 scenarios of each of the 14 scenarios type of the nuPlan challenge. See Figure \ref{tab:val14}. %Val14 benchmark.

We compare MBAPPE's performance with other state-of-the-art method on the validation scenario of the Val14 benchmark \cite{pdm}. See Table \ref{tab:val14}. %Val14 benchmark.

\textbf{Baselines: }
Urban Driver \cite{urbandriver} utilizes PointNet \cite{pointnet} layers to process polyline and employs a MLP following a multi-head attention block to forecast the ego trajectory. GameFormer Planner \cite{gameformer} exploits a Transformer to predict all agents trajectories before refining ego planning via non-linear optimization. PlanCNN \cite{plancnn} leverages a CNN on rasterized inputs to predicts the ego trajectory. PDM \cite{pdm} leverages an improved IDM \cite{idm} model combined with a simple MLP to generate several trajectories which are then scored to return the optimal one. GC-PGP \cite{gc-pgp} categorizes proposed plans according to their traversal of a route-constrained lane graph, and then identifies the most probable cluster center. 

\begin{table}[!ht]
    \begin{center}
        \resizebox{\columnwidth}{!}{
            \begin{tabular}{l|ccc|c}
                \toprule 
                Method & CR $\downarrow$ & DA $\downarrow$ & EP $\uparrow$  & Score $\uparrow$\\
                \midrule 

                Urban Driver MA \cite{urbandriver} & 34\% & 26\% & 96\%  & 47\%\\
                GameFormer Planner \cite{gameformer} & 6\% & 4\% & 98\% & 84\% \\
                PDM-Hybrid \cite{pdm} & \textbf{2\%} & \textbf{0\%} & \textbf{99\%} & \textbf{93\%}\\
                IDM \cite{idm} & 12\% & 6\% & 95\% & 76\%\\
                %Log Replay (GT) & 1\% & 1\% & 100\%  & 95\%\\
                GC-PGP \cite{gc-pgp} & - & - & - & 57\%\\
                PlanCNN \cite{plancnn} & - & - & - & 73\%\\
                %Urban Driver \cite{urbandriver} & - & - & -  & 44\%\\

                \midrule 
                MBAPPE (GameFormer) & 3\% &  2\% & 98\% & 90\%\\
                MBAPPE (Urban Driver) & 5\% & 2\% & 96\% & 86\%\\
                \bottomrule
            \end{tabular}
        }
    \end{center}
\caption{\textbf{Val14 benchmark on nuPlan}}
\label{tab:val14}
\end{table}

\begin{figure*}[!ht]
    \centering
    \includegraphics[width=\linewidth]{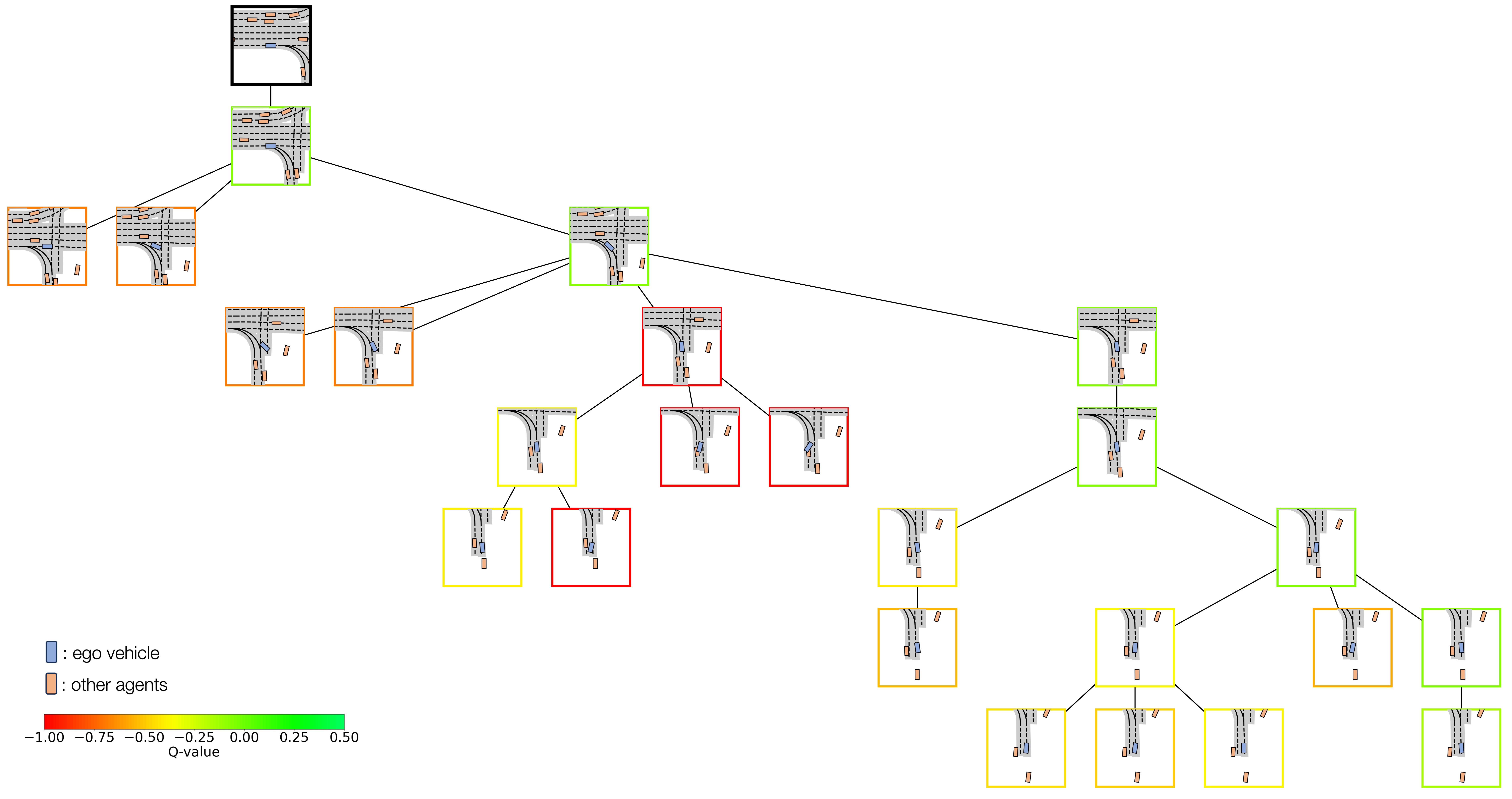}
    \caption{A subset of a decision tree obtained with MCTS exploration. Nodes are colored according to their Q-value. The root node correspond to the present state of the vehicle in the nuPlan simulator. We observe that the orange left branch exploration leads to the ego leaving the expected route, hence the low Q-value. The red middle branch exploration leads to a collision, thus explaining the low Q-value. The green right branch exploration presents the expected behavior and therefore has the highest Q-value. The explored planning can also be observed in Figure \ref{fig:exploration}.}
    \label{fig:decision_tree}
\end{figure*}

For this comparison, we extended Urban Driver to predict trajectories of all other agents in the scene in addition to the ego's. We name this updated version Urban Driver Multi-Agent (Urban Driver MA). Then, we evaluated two versions of MBAPPE. One leverages Urban Driver MA as prediction and prior model (c.f. Figure \ref{Fig:mbappe_fig}), and the other a GameFormer model. Other components of those systems are identical. 

In our experiments, we found that enhancing a prediction model with MBAPPE consistently results in improved planning. Specifically, when integrated with GameFormer, MBAPPE yields a substantial improvement in key metrics compared to using non-linear optimization techniques as done with the GameFormer Planner.

Thus, MBAPPE not only delivers state-of-the-art performance, but is also an explainable and interpretable operator when applied to predictive models. This dual benefit both refines decision-making policies and provides added adaptability.

\section{An explicit and explainable method}
%One of the main advantage of this method is that with very simple instructions incorporated in a reward function (go forward, don't collide, stay on the route, stay on the road) and a pretty vague prior we obtain a very realistic working planning, without resorting to ungeneralizable handcrafted rules (such as using the curvature of the current line, or follow the speed of the car in front) or to black boxes not explainable neural networks. This gives our approach plenty of flexibility, adaptability and explainability.

A key benefit of this technique is its simplicity: it requires only basic high-level directives in the form of a reward function (e.g., move ahead, avoid collisions, stick to the route, and remain on the road). Despite its vague prior, the method yields highly effective and realistic planning. This eliminates the need for specific, hard-to-generalize rules, like basing decisions on the road's curvature or the speed of the car ahead, as well as the use of hardly interpretable neural networks. As a result, our approach is highly flexible, adaptable, and explainable.

Indeed, decisions of the MCTS are explainable and the internal process that led to those decisions can be easily observed and analyzed. Figure \ref{fig:decision_tree} provides an example of a decision tree of the MCTS in which we can observe several exploration branches and their consequences on the tree expansion. In particular, we observe on the green right branch that internal exploration leading to desirable behavior yields the highest Q-value and further exploration of that branch. When exploration leads to collisions or to the ego leaving its expected route, the Q-value is low and exploration stops, as shown in the red middle and orange left branches. Figure \ref{fig:decision_tree} shows that MCTS decisions-making process is transparent and explainable, thus leading to an explicit and safe planning.

%AI for autonomous driving is not just a matter of science, but also a matter of safety. Therefore, we believe a broadly adopted AI system for autonomous driving should be 1) reliable, 2) explainable and 3) predictable. Hence, the insights given by the internal exploration of the MCTS provide a considerable step forward a safe autonomous driving system, while taking profit from the generalizability of supervised neural networks. 

%Our work process during the development of this solution consisted in visualizing the thought process of the MCTS in scenarios that resulted in failures and try to understand whether these were due to implementation bugs or poor decision making. With these informations, we were able to iteratively improve our solution and make tremendous progress towards our final performance. We believe that much is still to be done and achieve in terms of potential in this method. Such a process was only made possible with the interpretable nature of the MCTS way of doing exploration and taking decisions.

\section{Conclusion}

% This paper introduces MBAPPE, a novel method extending MCTS to partially learned environment for planning in an autonomous driving setup. We showed the added value of the designed priors and continuity constraints in the MCTS tree through ablation studies. We demonstrated that MBAPPE can be used as an improvement operator over prediction models via comparison on a nuPlan simulator benchmark as it outperforms vanilla models in each metrics for all experiments. Lastly, we insist of the interpretability this method offers, which is an important feature for safe and reliable autonomous driving.

%It then compared our method to state-of-the-art research through a benchmark on the nuPlan simulator to demonstrate that MCTS can be used as an improvement operator over prediction models. It finally emphasized on the interpretable nature of this method, therefore allowing an explicit planning both explainable and reliable.

This paper presents MBAPPE, a novel approach extending MCTS for planning within a partially learned environment in the context of autonomous driving. Through ablation studies, we highlighted the advantages of incorporating the designed priors and continuity constraints into the MCTS tree. Comparative analysis using a benchmark on the nuPlan simulator revealed that MBAPPE is an effective refinement operator for planning models, consistently outperforming vanilla models across all evaluation metrics. Finally, we emphasize the interpretability provided by this technique, a critical attribute for ensuring the safety and reliability of autonomous vehicles.

In terms of future work, as MBAPPE improves planning model capabilities, one could fine-tune the prior network similarly to the approach used in AlphaGo \cite{alphago}. This would enable the network to better emulate the MCTS output, thereby refining its priors and initiating a cycle of self-improvement. Better results could also be achieved with a more complex learned prior inferred for each node \cite{muzero, chen}, as well as learning a bootstrapped value network to estimate node expected returns in addition to the current reward. However this would require more network inferences and could harm the execution time.

\clearpage
% \bibliography{references}
%\bibliographystyle{IEEEtran}
\printbibliography

\end{document}